\documentclass[]{x2lab}

\usepackage{microtype}
\usepackage{graphicx}
\usepackage{subcaption}
\usepackage{tikz}
\usetikzlibrary{arrows.meta,positioning,calc,decorations.pathreplacing,patterns}
\usepackage{csquotes}
\usepackage{afterpage}
\usepackage{placeins}
\usepackage{float}

\usepackage{amsmath}
\usepackage{amssymb}
\usepackage{mathtools}
\usepackage{amsthm}
\usepackage{xspace}
\usepackage{colortbl}
\usepackage{multirow}
\usepackage{enumitem}
\usepackage{wrapfig}
\usepackage{nicefrac}
\usepackage{makecell}
\usepackage{xcolor}
\usepackage{pifont}

\definecolor{fbApp}{HTML}{c8e7fa}
\definecolor{fbPurple3}{HTML}{f0ebf5}

\definecolor{citecolor}{HTML}{0071BC}
\definecolor{linkcolor}{HTML}{ED1C24}
\definecolor{efblue}{RGB}{0, 102, 204}

\providecommand{\planner}{\textsc{X-Planner}\xspace}

\makeatletter
\renewcommand\paragraph{\@startsection{paragraph}{4}{\z@}%
  {1.5ex \@plus 1ex \@minus .2ex}%
  {-1em}%
  {\normalfont\normalsize\bfseries}}
\makeatother

\title{X-Planner: Event-Structured Task Planning\\for Embodied Intelligence}
\author{X Square Robot Team}

\abstract{
Task planning bridges high-level instructions and executable behavior in long-horizon manipulation, yet modern Vision--Language--Action (VLA) systems often leave this intermediate structure implicit. Existing chain-of-thought (CoT) planners also tend to rely on coarse task-level annotations or serialize long reasoning traces token by token. We present \planner, a planning front-end that addresses both the supervision and representation of embodied reasoning. Our planning data combine Ego, UMI, and teleoperation under a hierarchy granularity with source-dependent annotation depth. Takeover-time annotations and human-designed failures supervise ongoing error recognition. On the model side, a shared VLM backbone exposes two event-structured plan forms: a discrete interface that emits interpretable event states and a latent interface that relays continuous CoT states across staggered Transformer depths through \emph{Staircase Decoding}. A frozen latent-to-text reconstruction objective provides a semantic anchor for the latent representation. Offline two-step planning evaluation places \planner second among four evaluated models on both BERTScore-F1 and a judge-based Overall score. In real-robot experiments, respectively, outperforming the evaluated baselines. These results characterize planning-text quality and downstream execution.
}

\date{September 22, 2026}

\metadata[Code]{\url{https://github.com/X-Square-Robot/Xplanner}}

\begin{document}
\maketitle

\section{Introduction}
\label{sec:intro}

Embodied foundation models have advanced rapidly by mapping visual observations and natural-language instructions to executable actions~\cite{zitkovich2023rt,kim2024openvla,black2024pi_0,black2025pi_,bjorck2025gr00t,cheang2025gr}. This direct observation-to-action paradigm is effective for short, well-specified skills. Long-horizon tasks, however, require an agent to select and order grounded sub-goals before issuing low-level commands. Many Vision--Language--Action (VLA) models leave this intermediate structure implicit. Bridging high-level intent and low-level control is therefore a planning problem as well as a multimodal-representation problem. Recent systems begin to address it by introducing chain-of-thought (CoT) reasoning before action generation~\cite{zawalski2024embodiedcot,zhao2025cot}.

Despite this progress, embodied task planning remains fragmented across data construction, planning granularity, and decoding strategy. We focus on three gaps.

\paragraph{The data gap.}
Embodied planners are often trained with task-level descriptions or hand-written sub-goal lists for a fixed suite. Such labels rarely specify \emph{where} one sub-goal ends and the next begins. A single episode caption can obscure the internal event structure of a demonstration: regrasps, failed contacts, retries, and small pose corrections may all disappear behind a label such as ``pick up the object''~\cite{shou2021generic}. Planning supervision must therefore capture event boundaries and execution errors as they arise. Failure examples collected from a single policy may also overrepresent that policy's characteristic mistakes.

\paragraph{The granularity gap.}
Reasoning-based planners also tend to commit to a single, externally imposed granularity. Some emit text at a fixed semantic level, whereas others forecast sub-goal images or optical flow~\cite{zhao2025cot}. We instead adopt the \emph{semantic event} as the planning unit: a temporally coherent span of executable behavior, such as reaching, grasping, lifting, or placing, whose boundary follows a change in behavior. In contrast, a fixed-length chunk may split one behavior or merge several behaviors into a single target. Action-grounded events provide a useful middle ground: each plan step is meaningful in language, observable in video, and realizable through control.

\paragraph{The latency gap.}
Finally, conventional CoT planners decode reasoning autoregressively, one token at a time. Over a long rollout, this serial dependency can repeatedly process overlapping visual--language context and delay the next planning handoff~\cite{hao2024coconut,zhong2026dualcot}. Long explicit traces are also vulnerable to error propagation. Latent-CoT methods replace discrete tokens with continuous states, but many retain a serial dependency between latent steps. A practical planner should reduce this serial critical path while exposing a representation that the execution stack can consume at a well-defined handoff.

We present \planner, a task-planning front-end that addresses these three gaps jointly. Given a high-level instruction and the current multi-view observation, \planner produces an event-structured representation that conditions a downstream world-action model. The architecture is detailed in Sec.~\ref{sec:method} and Fig.~\ref{fig:planner-overview}. Our contributions are threefold.

\begin{enumerate}
    \item \textbf{Multisource, hierarchical planning supervision.} We combine Ego, UMI, and teleoperation under an L3 Task/L2 Subtask/L1 Action/L0 Segment hierarchy. Ego and UMI retain L1--L3; teleoperation supports all four levels. Annotated takeover times supervise ongoing error recognition, while human-designed failure demonstrations supplement policy-collected errors to reduce dependence on a single policy's failure distribution. A $1{,}500$-episode analysis subset characterizes semantic and temporal coverage.

    \item \textbf{Staircase Decoding with discrete and latent plan forms.} A shared VLM backbone produces either explicit event states or continuous CoT states relayed across staggered Transformer depths. The latent form avoids token-by-token serialization within the planner, while a frozen latent-to-text objective encourages the compact states to retain plan semantics.

    \item \textbf{Offline planning and real-robot evaluation.} Offline two-step text evaluation measures semantic matching and overall plan quality: \planner scores above Qwen and Doubao and below kimi3 on both reported metrics. Coupling \planner to a world-action backbone yields the highest average Task Progress among the evaluated systems on the reported reasoning and generalization suites.
\end{enumerate}

Together, these components frame embodied planning around action-grounded events, reproducible supervision, and complementary explicit and latent interfaces. The reported experiments evaluate planning text offline and the complete event-mode system on real robots; controlled comparisons of the two plan forms are left for future work.
\section{Related Work}
\label{sec:related}
\subsection{Reasoning and Planning in VLA Models}
Vision--Language--Action models extend pretrained vision--language models with interfaces that map observations and instructions to continuous control commands~\cite{zitkovich2023rt,kim2024openvla,octo2024,black2024pi_0,black2025pi_,bjorck2025gr00t,cheang2025gr}. Their semantic priors support generalization across objects, scenes, and instructions, yet the action interface is often trained as a largely reactive observation-to-action map with no explicit representation of task structure~\cite{li2024cogact,cheang2025gr}. A growing line of work introduces CoT reasoning to decompose a task before acting~\cite{zawalski2024embodiedcot,zhao2025cot,huang2025ladi}. Linguistic approaches emit textual sub-goals or reasoning traces, as in Embodied CoT~\cite{zawalski2024embodiedcot}; visual approaches forecast sub-goal images or dense motion cues, as in CoT-VLA~\cite{zhao2025cot}. These representations trade semantic readability against spatial specificity and generation cost. \planner instead grounds both its explicit and latent representations in action-aligned semantic events, leaving fine spatiotemporal realization to the downstream world-action model.

\subsection{Explicit, Latent, and Parallel Reasoning}
A complementary line of work studies \emph{how} reasoning is decoded. Explicit CoT generates discrete tokens sequentially, creating a serial critical path and allowing an early error to affect later steps. Latent-reasoning methods reduce textual serialization by routing CoT through compact continuous states~\cite{hao2024coconut,goyal2024think,kang2025ladir,kang2026beyond,zhong2026dualcot}. Coconut~\cite{hao2024coconut} and LaDiR~\cite{kang2025ladir}, for example, compress intermediate thoughts into continuous representations; recent embodied variants distill spatiotemporal or world-model foresight for manipulation and driving~\cite{liu2026last,luo2026last,bai2026latent}. Many such methods nevertheless retain an autoregressive dependency between latent steps. \planner's Staircase Decoding instead relays hidden states across staggered Transformer depths, allowing latent plan states to share lower-layer grounding computation and proceed in parallel through the upper layers.

\subsection{Data and Annotation for Embodied Planning}
Internet video~\cite{nan2024openvid} and egocentric corpora~\cite{grauman2022ego4d} provide visual-dynamics priors. Robot datasets~\cite{khazatsky2024droid,bu2025agibot} pair demonstrations with control actions. Episode-level task strings alone conceal the sub-goal structure required by a planner. Temporally grounded captioning and atomic-action segmentation~\cite{shou2021generic,hu2026finevla} motivate finer supervision. Our planning data (Sec.~\ref{sec:platform}) combine Ego, UMI, and teleoperation with source-dependent annotation depth, grounding event order and progress in a shared hierarchy. Takeover-time annotations and human-designed failures extend this supervision to recognizing errors during ongoing execution.
\section{Event-Structured Planning Data}
\label{sec:platform}

\subsection{Sources and Hierarchical Annotation}
\label{sec:platform-caption}

\begin{figure}[!htbp]
    \centering
    \includegraphics[width=\linewidth]{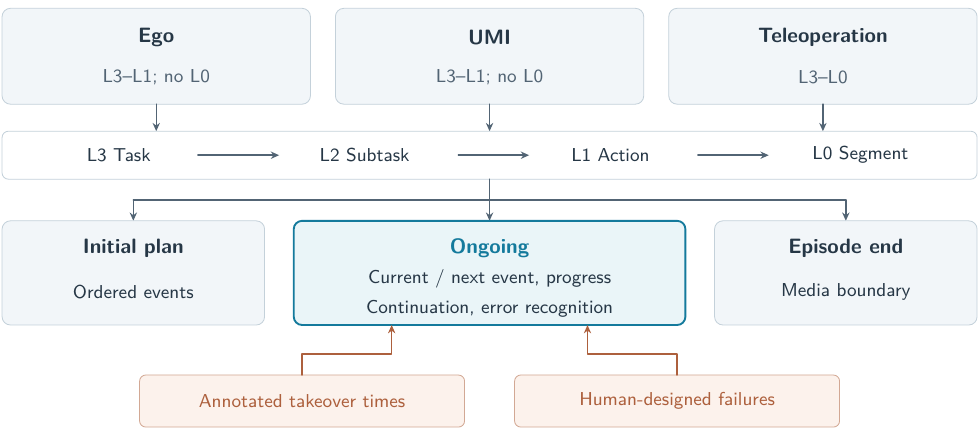}
    \caption{\textbf{Planning data and supervision.}
    Three data sources share a hierarchy with source-dependent granularity. Takeover timing and human-designed failures provide ongoing error-recognition supervision.}
    \label{fig:pipeline}
\end{figure}

We build planning supervision from egocentric (Ego), UMI, and teleoperated demonstrations (Fig.~\ref{fig:pipeline}). A shared hierarchy represents \textbf{L3 Task} (episode objective), \textbf{L2 Subtask} (semantic stage), \textbf{L1 Action} (executable primitive), and \textbf{L0 Segment} (finest temporal span). Teleoperation data carry all four levels. Ego and UMI retain L1--L3: their faster motions make L0 boundaries difficult to annotate reliably, so Action is their finest labeled level. Available visual and action streams are synchronized and screened for missing observations, timestamp inconsistencies, and unusable records before annotation.

\subsection{Failure-Aware Ongoing Supervision}
\label{sec:platform-ongoing}

Ongoing supervision must identify execution errors as well as track plan progress. We therefore annotate takeover times in intervention episodes, providing temporal supervision for error recognition during execution. Because these episodes reflect the failure distribution of the collecting policy, we supplement them with human-performed demonstrations of deliberately designed failure modes. These simulated failures broaden error coverage and are intended to reduce dependence on a single policy's biases. Together with the hierarchy, the annotations support initial-plan, ongoing, and episode-end states; ongoing targets cover current/next events, normalized progress, continuation state, and error recognition.

\subsection{Coverage and Selection}
\label{sec:platform-data-coverage}

A deterministic analysis subset retains $1{,}500$ of $1{,}654$ candidate episodes. Removing $154$ redundant, frequent combinations preserves all $167$ datasets, $30$ named task categories plus an unknown label, and the observed tail of semantic and action labels. Fixed-seed tie-breaking makes selection repeatable. These statistics describe the analyzed pool, rather than the full three-source training mixture or its source proportions.

\begin{figure}[!htbp]
    \centering
    \includegraphics[width=\linewidth]{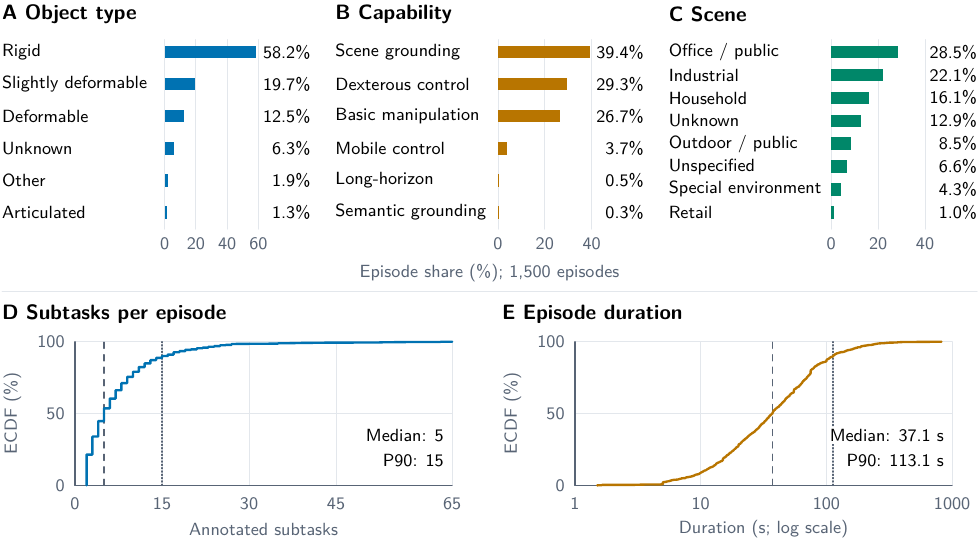}
    \caption{\textbf{Semantic and temporal coverage of the $1{,}500$-episode analysis subset.}}
    \label{fig:data-coverage}
\end{figure}

Figure~\ref{fig:data-coverage}A--C shows independent object, capability, and scene shares, dominated by rigid objects ($58.2\%$), scene grounding ($39.4\%$), and office/public scenes ($28.5\%$). The temporal empirical cumulative distribution functions (ECDFs; D--E) have medians of five subtasks and $37.1$~s, and $90$th percentiles of $15$ subtasks and $113.1$~s (dashed and dotted guides, respectively). Subtask counts reach $65$; episode duration is the maximum video duration across cameras, reaches $822.7$~s, and uses a logarithmic axis.

\begin{figure}[!htbp]
    \centering
    \includegraphics[width=\linewidth]{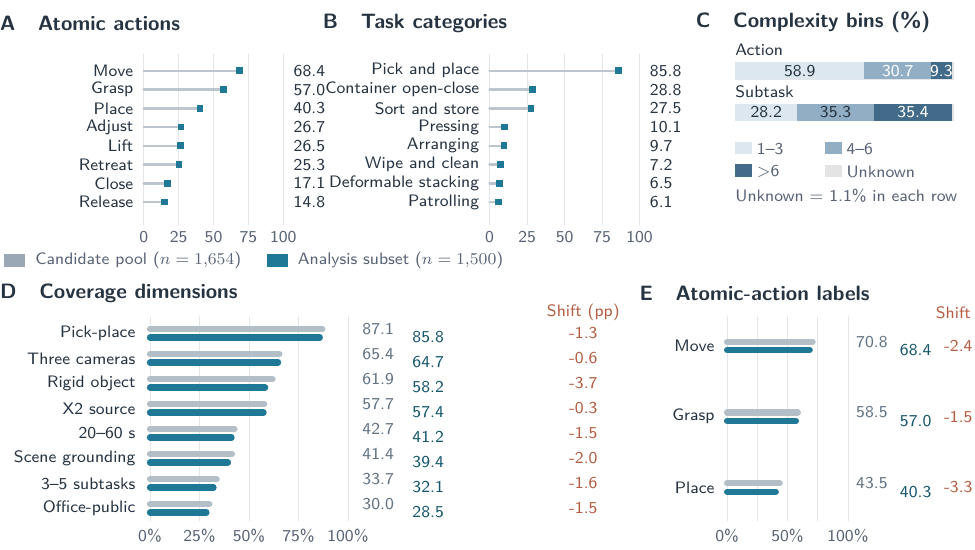}
    \caption{\textbf{Action/task profiles and selection effects.} A--C show episode shares and complexity-bin composition; D--E show retained-minus-candidate shifts (percentage points). All panels use the same $1{,}500$-episode subset and denominator; labels are multi-label where applicable.}
    \label{fig:data-coverage-actions-tasks}
    \label{fig:data-coverage-complexity}
    \label{fig:data-coverage-selection}
\end{figure}

Figure~\ref{fig:data-coverage-actions-tasks}a--b characterizes action and task coverage. The manifest contains $41$ named atomic-action labels inferred from two ground-truth subtask captions per episode, with one unmatched episode. Move, grasp, and place cover $68.4\%$, $57.0\%$, and $40.3\%$ of episodes; pick-and-place, opening/closing containers, and sorting/storage cover $85.8\%$, $28.8\%$, and $27.5\%$. These are multi-label episode shares, not action-occurrence frequencies. All $1{,}500$ episodes remain in the denominator, including $17$ with an unknown task label; the complete counts preserve the source taxonomy.

Figure~\ref{fig:data-coverage-complexity}c contrasts source-defined action- and subtask-complexity metadata. The action bins $1$--$3$, $4$--$6$, and $>6$ account for $58.9\%$, $30.7\%$, and $9.3\%$, respectively. Subtask-complexity shares are $28.2\%$, $35.3\%$, and $35.4\%$; both metadata fields have $1.1\%$ unknown labels. These categories remain separate from measured subtask counts and caption-inferred action types and do not constitute calibrated difficulty scores.

Figure~\ref{fig:data-coverage-selection}D--E reports the effect of deterministic selection as the retained share minus the candidate-pool share. Every displayed shift is negative because the procedure removes redundant frequent combinations instead of reweighting labels: the largest coverage change is for rigid objects ($-3.7$ percentage points), followed by scene grounding ($-2.0$~pp), while the source-group and camera-count shifts are at most $-0.6$~pp. The same pattern appears for atomic actions, with Place ($-3.3$~pp) and Move ($-2.4$~pp) changing most and Grasp changing by $-1.5$~pp. Thus, selection trims dominant modes while keeping the taxonomy and long tail available; these are episode-share shifts, not changes in per-action frequency or calibrated difficulty.

\FloatBarrier
\section{Method}
\label{sec:method}

\begin{figure}[!htbp]
    \centering
    \includegraphics[width=\linewidth]{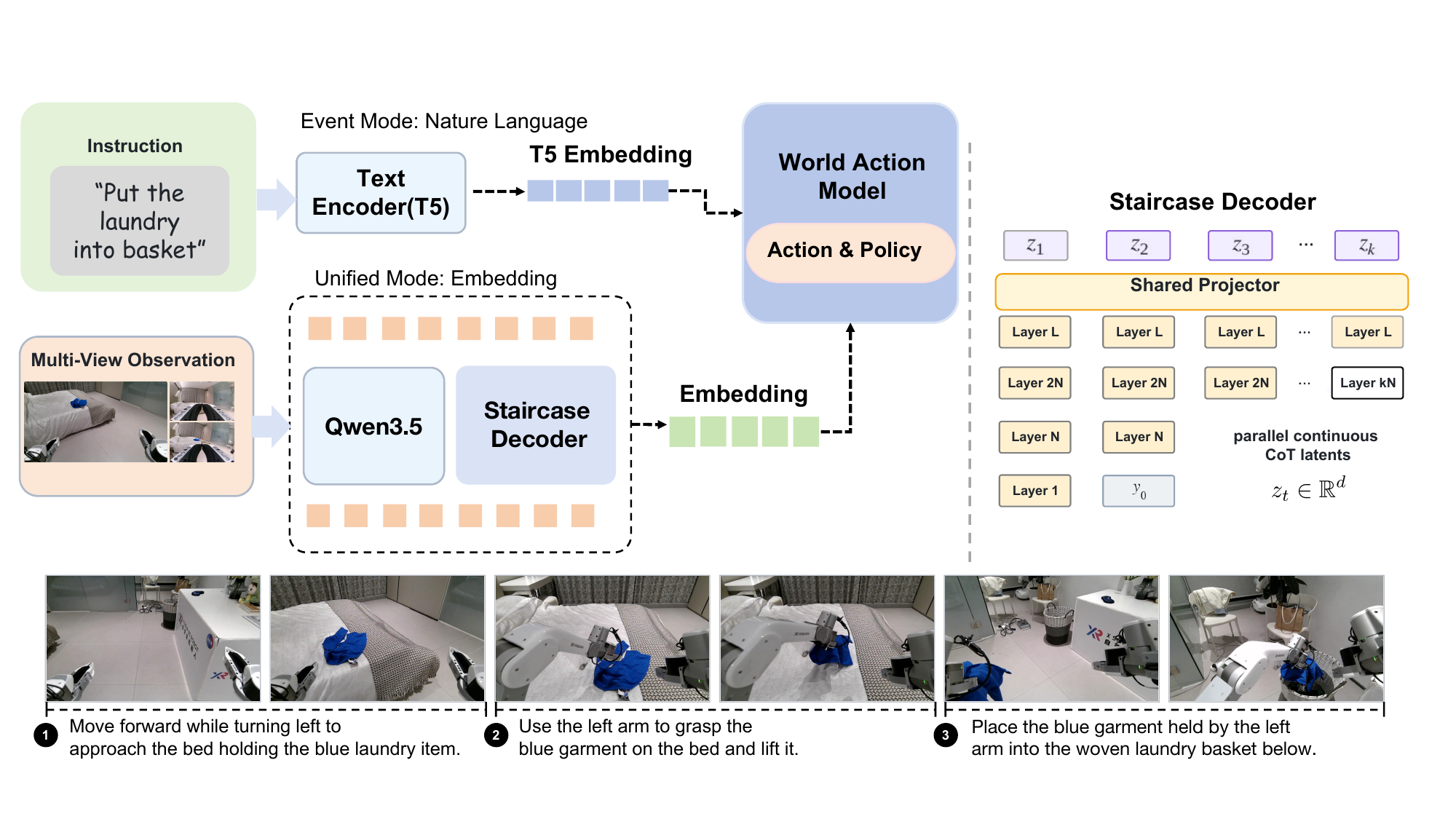}
    \caption{\textbf{Overview of \planner.}
    Given a high-level instruction and multi-view observations, the planner exposes an interpretable event-language interface and a continuous conditioning interface to a downstream world-action model. Staircase Decoding constructs latent plan states across staggered Transformer depths; the lower row illustrates an event-level rollout.}
    \label{fig:planner-overview}
\end{figure}

Given a high-level instruction $\ell$, the available visual observations $\{\mathbf V_0^{(v)}\}_v$, and optional history $h$, \planner produces an event-structured representation that conditions a downstream world-action model. The view index $v$ ranges over the views available in each example. A Qwen-series VLM backbone~\cite{qwen25,qwen35} provides the visual--language features used for scene grounding, event decomposition, and progress estimation. As shown in Fig.~\ref{fig:planner-overview}, the architecture exposes the plan in two forms: a discrete event state for interpretable planning and a latent state sequence that avoids token-by-token serialization within the planner.

\subsection{Two Plan Forms}
\label{sec:method-forms}

\paragraph{Discrete form: structured event states.}
In the discrete form, the VLM emits a compact event state rather than an isolated caption. At the start of execution, it produces an ordered initial plan. During rollout, its structured response covers normalized progress, the current event, the next event when available, continuation state, and execution-error recognition. An event caption may, for example, read \textit{align the gripper above the red cup}. At the final available media frame, the model emits an episode-end marker. These structured states are inspectable and editable by a human or upstream agent, and their event descriptions provide the text interface to the downstream world-action model. The episode-end marker denotes the recorded-media boundary; it is not, by itself, a certificate of semantic task success.

\paragraph{Latent form: parallel continuous CoT.}
The discrete form relies on autoregressive token generation. The latent form instead represents the plan as a compact sequence of $K_c$ continuous reasoning states, $\hat y_{1:K_c}=\{\hat y_1,\ldots,\hat y_{K_c}\}$. The staircase branch generates these states in parallel and injects them directly into the downstream cross-attention pathway. Both forms share the VLM backbone and event-grounded semantics (Sec.~\ref{sec:training}); they differ in whether the plan is exposed as text or retained as continuous states.

\subsection{Text Conditioning Interface}
\label{sec:method-text}
Both plan forms use a common conditioning interface. Let $\mathbf H_q$ denote the VLM hidden states. A boundary mask $\mathbf m_{\texttt{txt}}$ selects the text-token states, which are projected into the downstream conditioning space:
\begin{equation}
\mathbf c_\ell
=
\mathcal M_T\!\bigl(\mathcal P_Q(\mathbf H_q[\mathbf m_{\texttt{txt}}])\bigr),
\end{equation}
where $\mathcal P_Q$ is a learnable projection and $\mathcal M_T$ is the downstream backbone's native text MLP. The resulting sequence enters the world-action cross-attention layers alongside image tokens. The alignment objective in Sec.~\ref{sec:training-alignment} encourages these projected VLM states to match the text-feature geometry expected by the fixed downstream model. This interface allows the planner to add scene-grounded disambiguation, task decomposition, and progress-aware event context without replacing the downstream text pathway.

\subsection{Staircase Latent CoT}
\label{sec:method-staircase}
The latent form is implemented through \emph{Staircase Decoding}. In a conventional autoregressive latent rollout, each reasoning state depends on the previous state and traverses the full Transformer stack. Staircase Decoding shortens this serial path through a depth-parallel schedule.

The reasoning branch is initialized from the fine-tuned VLM and implemented as a lightweight Mixture-of-Transformers (MoT) structure coupled to the frozen backbone. We partition the Transformer at a \emph{relay depth} $N_r$. The lower layers encode shared visual--language context, whereas the upper layers specialize across reasoning steps. The first latent position traverses the lower layers to produce a relay representation shared by all reasoning positions. The upper blocks then update the latent states in parallel, with an independent causal cache for each position:
\begin{equation}
\hat y_{1:K_c} = \mathcal F_{\mathrm{stair}}(x; N_r),
\end{equation}
where $x=(\{\mathbf V_0^{(v)}\}_v,\ell,h)$ and $N_r$ controls how much grounding computation is shared before the reasoning paths diverge. Relative to a fully autoregressive rollout, the schedule reuses lower-layer visual--language features across reasoning states (Fig.~\ref{fig:planner-overview}). The resulting latents condition the downstream cross-attention pathway without discrete token sampling. This architectural reduction in serial computation does not, by itself, establish lower end-to-end control latency, which also depends on the surrounding execution stack.

\subsection{Frozen Latent-to-Text Reconstruction}
\label{sec:method-recon}
A latent plan is useful only if its compact states retain the intended event semantics. We encourage this property with a \emph{frozen latent-to-text reconstruction} objective~\cite{kang2025ladir,zhong2026dualcot}, rather than by matching a particular sequence of autoregressive hidden states. A prefix projector $\mathcal P_{\mathrm{pref}}$ maps $\hat y_{1:K_c}$ to a soft prefix $\mathbf z_{1:K_c}=\mathcal P_{\mathrm{pref}}(\hat y_{1:K_c})$ in the embedding space of a lightweight frozen language model. Conditioned on this prefix, the language model reconstructs the corresponding textual CoT trace:
\begin{equation}
P_\phi(r_{1:M_r}\mid \mathbf z_{1:K_c}),
\qquad
\mathcal L_{\mathrm{CoT}}
=
-\sum_{m=1}^{M_r}\log P_\phi\!\left(r_m \mid \mathbf z_{1:K_c}, r_{<m}\right).
\end{equation}
Only the staircase reasoning branch and $\mathcal P_{\mathrm{pref}}$ are optimized; the reconstruction model remains fixed. Reconstructing the trace encourages the latent sequence to preserve high-level semantics without copying a specific token-level hidden trajectory. The frozen decoder can also render the latent plan as text for qualitative inspection.

\subsection{Inference: Coupling \planner to a World-Action Model}
\label{sec:method-inference}
The two plan forms define corresponding deployment modes. In \textbf{event mode}, \planner emits an initial plan at startup and updates its structured event state as new observations arrive. Event descriptions condition the world-action model, and ongoing error recognition supports execution monitoring. The execution system determines when to refresh or commit the state. In \textbf{unified mode}, the staircase decoder produces $K_c$ continuous CoT states that condition fixed-length action-chunk inference through the same cross-attention interface. The common interface therefore supports either readable event plans or continuous latent conditioning.
\section{Training}
\label{sec:training}

Each training example combines the available visual observations, a task instruction, optional history, and a complete compact JSON target. The discrete interface is trained on this structured response, while feature alignment and staircase reconstruction introduce separate model-side objectives. The downstream world-action model remains fixed throughout.

\subsection{Structured Plan-State Supervision}
\label{sec:training-text}
The data pipeline materializes three event-state categories before optimization (Table~\ref{tab:train-states}). Supervision uses only the annotation levels supported by each source: Ego and UMI provide L1--L3, while teleoperation additionally provides L0. For failure examples, annotated takeover times and human-designed failures supervise error recognition during ongoing execution.

\begin{table}[!htbp]
  \centering
  \small
  \caption{Structured states used for discrete planning supervision.}
  \label{tab:train-states}
  \begin{tabular}{@{}>{\raggedright\arraybackslash}p{0.18\linewidth}>{\raggedright\arraybackslash}p{0.25\linewidth}>{\raggedright\arraybackslash}p{0.45\linewidth}@{}}
    \toprule
    State & Visual anchor & Supervised assistant state \\
    \midrule
    Initial plan & Execution start & Ordered initial event plan \\
    Ongoing & During execution & Current/next events, normalized progress, continuation state, and error recognition on annotated failure examples \\
    Episode end & Last available media frame & Final event state and an episode-end marker \\
    \bottomrule
  \end{tabular}
\end{table}

Let $x=(\{\mathbf V^{(v)}\}_v,\ell,h)$ denote the observations, instruction, and optional history, and let $y_{1:M}$ denote the complete assistant JSON target. We supervise every assistant token autoregressively, rather than optimizing only the next-event caption or regressing progress as an independent scalar:
\begin{equation}
  \mathcal{L}_{\mathrm{SFT}}
  =
  -\sum_{m\in\mathcal A}\log p_\theta\!\left(y_m\mid x,y_{<m}\right),
\end{equation}
where $\mathcal A$ indexes assistant positions. Captions, progress, continuation state, and annotated errors are learned jointly within the structured response. Missing annotation levels contribute no supervision; in particular, the absence of L0 labels in Ego and UMI is not a negative Segment label.

\subsection{Feature Alignment for Text Conditioning}
\label{sec:training-alignment}
Separately from the JSON supervision, we align the planner-side conditioning path with the downstream model's original text-feature space:
\begin{equation}
  \mathcal{L}_{\mathrm{align}}
  =
  \lambda_{\mathrm{align}}\,\bigl\lVert\mathbf c_\ell^{\mathrm{VLM}}-\mathbf c_\ell^{\mathrm{ref}}\bigr\rVert^2,
\end{equation}
where $\mathbf c_\ell^{\mathrm{ref}}$ is the original text encoder's representation of the same instruction. This compatibility objective encourages the planner representation to remain in the geometry expected by the fixed downstream interface. It is a model-side loss, not an alternative assistant target.

\subsection{Staircase Distillation}
\label{sec:training-staircase}
The staircase module of Sec.~\ref{sec:method-staircase} is a lightweight MoT branch coupled to the frozen VLM backbone. Initializing the branch from pretrained VLM weights transfers the backbone's visual--language representations to the parallel latent rollout. Given the multimodal input, the branch produces $\hat y_{1:K_c}$, which serves as the implicit plan supplied to downstream cross-attention.

For supervision, the prefix projector $\mathcal P_{\mathrm{pref}}$ maps the latent sequence to a soft prefix $\mathbf z_{1:K_c}$ in the embedding space of a frozen lightweight language model. The language model then reconstructs the textual CoT trace autoregressively:
\begin{equation}
P_\phi(r_{1:M_r}\mid \mathbf z_{1:K_c}).
\end{equation}
The latent-to-text objective $\mathcal L_{\mathrm{CoT}}$ from Sec.~\ref{sec:method-recon} updates only the staircase branch and $\mathcal P_{\mathrm{pref}}$; all other components remain frozen. The targets $r_{1:M_r}$ are annotation-derived traces that share the event semantics of the discrete form. This reconstruction stage is distinct from compact-JSON SFT: the two paths share the backbone and semantic grounding, but use different targets and readouts.

\paragraph{Scope of the objectives.}
The structured assistant response defines the SFT target, feature alignment regularizes the downstream interface, and latent reconstruction supervises the staircase representation. We report these roles separately because the current system-level experiments do not isolate their individual causal contributions.
\section{Experiments}
\label{sec:exp}

We evaluate \planner at two complementary levels: the semantic and overall quality of two-step planning text, and the physical behavior of a \planner-conditioned world-action system. The real-robot evaluation covers reasoning beyond single-object motion and generalization under clutter and changing instructions.

\subsection{Offline Two-Step Planning Evaluation}
\label{sec:exp-offline}

The offline comparison evaluates two-step planning predictions against reference annotations. We report two complementary metrics: semantic matching of the predicted text and the overall quality of the proposed plan.

\paragraph{Metrics.}
\emph{BERTScore-F1}~\cite{zhang2020bertscore} measures semantic similarity between the predicted planning text and the reference annotations using contextual token representations. We report the F1 score without baseline rescaling. It measures text-level semantic agreement and should not be interpreted as a task-success rate. \emph{Overall} is a composite quality rating assigned by \emph{Doubao Judge}. Given the task description, reference answer, and model prediction, the judge considers accuracy, continuity, and the overall reasonableness of the planning result. Higher values are better for both metrics.

\begin{table}[!htbp]
    \centering
    \small
    \caption{\textbf{Offline two-step planning quality.}
    BERTScore-F1 is reported without baseline rescaling; Overall is the composite rating from Doubao Judge. Bold indicates the highest value in each column.}
    \label{tab:planning-text-results}
    \setlength{\tabcolsep}{14pt}
    \renewcommand{\arraystretch}{1.12}
    \begin{tabular}{lrr}
        \toprule
        Model & BERTScore-F1 $\uparrow$ & Overall $\uparrow$ \\
        \midrule
        Qwen & 0.8878 & 1.367 \\
        Doubao & 0.8846 & 1.366 \\
        kimi3 & \textbf{0.9108} & \textbf{1.434} \\
        \rowcolor{metabg}
        \planner (ours) & 0.9011 & 1.411 \\
        \bottomrule
    \end{tabular}
\end{table}

\paragraph{Results.}
Table~\ref{tab:planning-text-results} shows that \planner ranks second among the four evaluated models on both metrics. Its BERTScore-F1 of $0.9011$ exceeds Qwen and Doubao by $0.0133$ and $0.0165$, respectively, while remaining $0.0097$ below kimi3. The Overall rating follows the same ordering: \planner scores $1.411$, compared with $1.367$ for Qwen, $1.366$ for Doubao, and $1.434$ for kimi3. The two metrics thus give a consistent ranking in this comparison, with \planner producing stronger semantic matches and higher judge-rated plan quality than Qwen and Doubao. These aggregate scores characterize offline planning text; they do not establish statistical significance or physical execution success.

\FloatBarrier
\subsection{Real-Robot Evaluation}
\label{sec:exp-real-robot}

The real-robot comparison uses matched physical task definitions, language instructions, multi-view observations, scene-randomization protocols, and scoring rubrics.

\paragraph{Setup and metric.}
We conduct the experiments on tabletop bimanual arms using synchronized multi-view observations. The primary metric is \emph{Task Progress}, a dense $0$--$100$ score that credits partial completion according to task-specific rubrics. Unlike binary success, it captures partial grounding and execution progress. The evaluated \planner system operates in \emph{event mode}: given the task instruction and current observation, the planner produces a next-event description that conditions the world-action backbone. \textsc{U-Scratch} is a fixed-length instruction-to-action system trained from scratch with the same real-robot supervision. We treat it as a system-level reference rather than an intervention that isolates a particular \planner component. Where deployable, we additionally report $\pi_{0.5}$~\cite{black2025pi_}, DreamZero~\cite{ye2026world}, and LingBot-VA~\cite{li2026causal}.

\begin{figure*}[t]
    \centering
    \includegraphics[width=\textwidth]{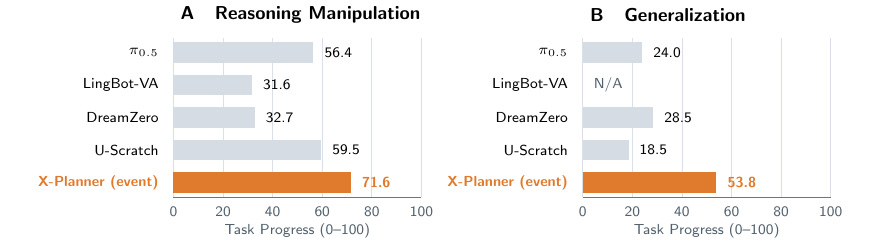}
    \caption{\textbf{Suite-level real-robot results.}
    Average Task Progress for (A) Reasoning Manipulation and (B) Generalization; higher is better. Each bar is the reported suite mean, and the value is printed at its endpoint. LingBot-VA is marked N/A in panel B because no result is available for that setting. Error bars are omitted because trial-level variation and uncertainty estimates were not provided.}
    \label{fig:real-robot-summary}
\end{figure*}

Figure~\ref{fig:real-robot-summary} consolidates the reported suite-level means in a single visual scale. Values are annotated directly to avoid requiring visual interpolation. LingBot-VA was not available in the Generalization setting and is marked \emph{N/A}. The available records contain neither trial-level scores nor a reported uncertainty statistic, so the figure intentionally shows no error bars; it should be read as a comparison of reported means rather than a variance or significance analysis.

\subsubsection{Reasoning Manipulation}
\label{sec:exp-reasoning}
The Reasoning Manipulation suite contains \texttt{Sort Headphone}, \texttt{Classify Items as Shape}, \texttt{Press Button in Order}, \texttt{Pair Up Items}, and \texttt{Pick Fruits into Basket}. Collectively, these tasks test category grounding, relational matching, ordered execution, and instruction-conditioned selection. Because each scene admits several plausible actions, the policy must identify the intended object, category, relation, or sequence before acting.

In Fig.~\ref{fig:real-robot-summary}A and Table~\ref{tab:reasoning-task-results}, the \planner-conditioned policy achieves the highest average Task Progress, $71.60$. The corresponding averages are $59.50$ for \textsc{U-Scratch}, $56.40$ for $\pi_{0.5}$, $32.70$ for DreamZero, and $31.60$ for LingBot-VA. Thus, \planner exceeds the strongest reported baseline in this suite, \textsc{U-Scratch}, by $12.10$ percentage points.

\begin{table}[!htbp]
    \centering
    \small
    \caption{\textbf{Task-level Reasoning Manipulation results.}
    Task Progress (\%); higher is better. Task values are rounded chart readouts; Average uses the separately reported suite mean.}
    \label{tab:reasoning-task-results}
    \setlength{\tabcolsep}{5pt}
    \renewcommand{\arraystretch}{1.04}
    \begin{tabular}{@{}lrrrrrr@{}}
        \toprule
        Model & \makecell[r]{Sort\\Headphone} & \makecell[r]{Classify Items\\as Shape} & \makecell[r]{Press Button\\in Order} & \makecell[r]{Pair Up\\Items} & \makecell[r]{Pick Fruits\\into Basket} & Average \\
        \midrule
        $\pi_{0.5}$ & 41\% & 55\% & 31\% & 77\% & 78\% & 56.40\% \\
        LingBot-VA & 56\% & 30\% & 8\% & 4\% & 60\% & 31.60\% \\
        DreamZero & 66\% & 36\% & 3\% & 13\% & 45\% & 32.70\% \\
        \textsc{U-Scratch} & 84\% & 82\% & 18\% & 43\% & 70\% & 59.50\% \\
        \rowcolor{metabg}
        \textbf{\planner} & 84\% & 78\% & 64\% & 36\% & 96\% & \textbf{71.60\%} \\
        \bottomrule
    \end{tabular}
\end{table}

Table~\ref{tab:reasoning-task-results} shows stronger Task Progress for \planner on ordered button pressing and instruction-conditioned fruit selection, with the same rounded score as \textsc{U-Scratch} on headphone sorting. \textsc{U-Scratch} scores higher on shape classification, while $\pi_{0.5}$ performs best on item pairing. These task-level patterns are descriptive; the suite-level comparison uses the separately reported means.

\subsubsection{Generalization}
\label{sec:exp-gen}
The Generalization suite tests instruction-conditioned transfer in a shared, cluttered tabletop scene. Figure~\ref{fig:real-robot-summary}B reports average Task Progress of $53.75$ for \planner, $28.50$ for DreamZero, $24.00$ for $\pi_{0.5}$, and $18.50$ for \textsc{U-Scratch}. The margin over the strongest available baseline, DreamZero, is $25.25$ percentage points. LingBot-VA is excluded because no result is available for this suite.

\FloatBarrier
\subsection{Interpreting the Results}
\label{sec:exp-interpretation}
The offline text metrics and real-robot Task Progress address different questions: whether the planner produces semantically appropriate, coherent text, and whether the conditioned system makes physical task progress. \planner ranks second on both offline metrics and achieves the highest reported mean Task Progress on both robot suites. The robot results evaluate the complete event-mode system; they do not separately estimate the contribution of error-supervision data, the discrete plan representation, the latent form, the staircase schedule, or any individual training objective. Neither evaluation establishes an end-to-end latency advantage. Controlled ablations and timing measurements are required to answer those questions.

\paragraph{Summary.}
The two levels of evaluation characterize both planning-text quality and downstream execution. Together with the event-structured data in Sec.~\ref{sec:platform}, the results support event-grounded planning as a useful interface between high-level instructions and action generation, while leaving a measurable gap to kimi3 in the reported offline text comparison.

\FloatBarrier
\section{Discussion and Conclusion}
\label{sec:conclusion}

We introduced \planner, a task-planning front-end that connects event-grounded supervision to downstream action generation. Its Ego, UMI, and teleoperation data use a shared hierarchy with source-dependent annotation depth. Takeover-time annotations and human-designed failures supervise ongoing error recognition, with designed failures intended to reduce dependence on a single policy's error distribution. The shared VLM backbone exposes both explicit event descriptions and continuous latent CoT representations, with Staircase Decoding reducing serialization within the latent planner. Offline two-step planning evaluation yields BERTScore-F1 of $0.9011$ without baseline rescaling and a Doubao Judge Overall rating of $1.411$, placing \planner above Qwen and Doubao and below kimi3. In the reported real-robot evaluation, the complete event-mode system achieves the highest average Task Progress among the evaluated methods on both reasoning and generalization suites.

\paragraph{Events as the interface between control and agents.}
Per-event textual plans provide a natural bridge between high-level agents and embodied control. \planner organizes this interface around task instructions, event states, and downstream vision-action execution. The discrete form gives a human or upstream agent an interpretable representation for inspecting or revising a plan, while the downstream model remains responsible for realizing it physically.

\paragraph{Decoupling interpretability and computation.}
Embodied reasoning must balance interpretability, which favors explicit plans, against compact computation, which favors continuous representations. \planner exposes both interfaces: the discrete form provides readable event states, while latent-to-text reconstruction gives the continuous form a textual semantic anchor. The staircase schedule shortens the planner's serial decoding path, although any end-to-end latency benefit depends on the complete execution and scheduling stack.

\paragraph{Limitations and outlook.}
The evaluation covers offline planning text and tabletop bimanual execution with the complete event-mode system. Offline results are aggregate text and judge scores, and do not establish physical success or statistical significance. Training histories and optional initial plans are annotation-derived, whereas deployment conditions on rolling model outputs; rollout-aligned supervision is needed to narrow this oracle-context gap. Episode-end labels mark synchronized media boundaries and do not independently establish task success. Controlled comparisons of the two plan forms, end-to-end timing, and evaluation on mobile or humanoid platforms remain future work. A further direction is to replace offline event discovery with a streaming process that can revise plans online. Within these limits, \planner provides a modular planning interface for embodied foundation models.

\section*{Contributors and Acknowledgments}
Howard Lu\textsuperscript{*}, Shalfun Li\textsuperscript{*\dag}, Porter Pan\textsuperscript{*}, Cris\textsuperscript{*},
Lumen, Cyril, Eric Hu, Lily Li, Maeve Zhang, Rain Sun,Robert Wang, KZ Zheng, Viggo Chen, Tim Ding, Regsis Cheng, YJ Xiao, Kian, Hai Lin, Alan Song, Elise Ma, Gody Li, Victor Yao, Yohann Tang, Ingrid Yu, Jason He,
James Wang, Ryan Yu, Ping Yang, Chris Pan, Vincent Chen, Roy Gan,
Hao Wang\textsuperscript{\ddag}, Qian Wang.

\noindent\textsuperscript{*}Core contributor.\quad
\textsuperscript{\dag}Project lead.\quad
\textsuperscript{\ddag}Corresponding author.

\clearpage
\bibliographystyle{plainnat}
\bibliography{paper}

\end{document}